\documentclass[11pt,a4paper]{article}
\usepackage[hyperref]{emnlp}
\usepackage{times}
\usepackage{latexsym}

\usepackage{url}

\usepackage{graphicx}
\graphicspath{ {images/} }
\usepackage{booktabs}
\usepackage[colorinlistoftodos]{todonotes}

\aclfinalcopy 

\title{Firearms and Tigers are Dangerous, Kitchen Knives and Zebras are Not:\\Testing whether Word Embeddings Can Tell}

\author{Pia Sommerauer and  Antske Fokkens\\
  Computational Lexicology and Terminology Lab \\
  Vrije Universiteit Amsterdam \\
  De Boelelaan 1105 Amsterdam, The Netherlands \\
  {\tt pia.sommerauer@vu.nl,  antske.fokkens@vu.nl} \\}

\date{}

\begin{document}
\maketitle
\begin{abstract}
This paper presents an approach for investigating the nature of semantic information captured by word embeddings. We propose a method that extends an existing human-elicited semantic property dataset with gold negative examples using crowd judgments. 
Our experimental approach tests the ability of supervised classifiers to identify semantic features in word embedding vectors and compares this to a feature-identification method based on full vector cosine similarity. The idea behind this method is that properties identified by classifiers, but not through full vector comparison are captured by embeddings. Properties that cannot be identified by either method are not. Our results provide an initial indication that semantic properties relevant for the way entities interact (e.g.\ dangerous) are captured, while perceptual information (e.g.\ colors) is not represented. We conclude that, though preliminary, these results show that our method is suitable for identifying which properties are captured by embeddings.
\end{abstract}


\section{Introduction}

Word embeddings are widely used in NLP and have been shown to boost performance in a large selection of tasks ranging from morphological analysis to sentiment analysis \cite[among many others]{Lazaridou:2013,Socher:2013,Zhou:2015}. Despite a number of different approaches to evaluation, our understanding of what type of information is represented by the vectors remains limited. Most approaches focus on full-vector comparison which treat vectors as points in a space \citep{Yaghoobzadeh2016}, which are evaluated by performance on semantic similarity or relatedness test sets and analogy questions \citep{Mikolov2013,Turney2012}. 
Previous work, however, has shown that high performance does not necessarily mean that vectors actually contain the information required to solve the task \citep{Rogers2017,Linzen2016}. Better understanding of the kind of semantic information captured by word embeddings can increase our understanding of how they help improve downstream tasks. In general, understanding what information is present in (often prominent) input embeddings forms an essential component of gaining deeper understanding of the nature of information and manner in which it travels through the hidden layers of a neural network. 

In this paper, we propose a method that investigates what kind of semantic information is encoded in vectors using a human-elicited dataset of semantic properties. We compare the output of supervised classifiers to an approach based on full-vector comparison that cannot access individual dimensions. The assumptions behind this approach are that (1) both full-vector comparison and the supervised classifier will perform well on identifying semantic properties that correlate highly with general similarity; (2) the classifier will outperform full-vector analysis on properties that are reflected by the context, but shared among a diverse set of entities and (3) that neither approach will perform well on properties that are not represented directly or indirectly in the text. The last two outcomes can indicate whether a semantic property is encoded in embeddings (2) or not (3).

The main contribution of this paper lies in the new method and corpus it proposes. To our knowledge, this is the first approach that aims at identifying whether specific semantic properties are captured by individual dimensions or complex patterns in the vector. In addition, we provide specific hypotheses as to which properties are captured well by which method and test them using our approach.\footnote{The hypotheses and code for our experiments can be found at: \url{https://cltl.github.io/semantic_space_navigation}} Our general hypothesis states that semantic properties that are relevant for the way entities interact with the world are well represented (e.g.\ functions of objects, activities entities are frequently involved in), whereas properties of relatively little consequence for the way entities interact with the world are not (e.g.\ perceptual properties such as shapes and colors, which either have no function or highly diverse functions). Though preliminary due to the complexity of the task, results indicate that these tendencies hold. Moreover, the overall outcome shows that the method and data are complementary to existing intrinsic evaluation methods.

The rest of this paper is structured as follows. We discuss related work in Section~\ref{sec:relwork}. Our method is outlined in Section~\ref{sec:method}. Section~\ref{sec:experiments} presents our experiments and results. We finish with a critical discussion and overview of future work in Section~\ref{sec:discussion}. 

\section{Related work}\label{sec:relwork}

Intrinsic evaluation of word embeddings has primarily focused on two main tasks: identifying general semantic relatedness or similarity and the so-called analogy task, where word embeddings have been shown to be able to predict missing components of analogies of the type A is to B as C is to D \citep{Mikolov2013,Turney2012}. Furthermore, most intrinsic evaluation methods take full vectors into consideration. The famous examples $Paris - France + Italy \approx Rome$ or $king - man + woman \approx queen$ evoke the suggestion that embeddings can capture semantic properties. The task has, however, been criticized substantially \cite[among others]{Linzen2016,Gladkova2016a,Gladkova2016b,Drozd2016}.   

\citet{Gladkova2016b} follow an observation in \citet{Levy2014} on the large differences in performance on different categories in the Google analogy set \cite{Mikolov2013}. They provide a new, more challenging, analogy dataset that improves existing sets on balance (capturing more semantic categories) and size. \citet{Linzen2016} points out more fundamental problems including the observation that the target vector in the analogy task can often be found by simply taking the vector closest to the source. \citet{Drozd2016} show that classifiers picking out the target word from a set of related terms outperform the standardly applied cosine addition or multiplication methods. Though also boosted by the aforementioned proximity bias, these results indicate that standard methods of solving analogies miss information that is captured by embeddings. \citet{Rogers2017} conclude that the analogy evaluation does not reveal if word embedding representations indeed capture specific semantic properties. 

On top of that, an embedding may capture specific semantic properties in ways that are not analogous to semantic properties of related categories. Analogy methods assume that semantic properties stand in analogous relation to each other based on the information provided by the context, but there is no reason why (e.g.) things made of wood and things made of plastic result in (combinations of) embedding dimensions that are similar enough to stand in a parallel relation to each other. Our set-up can determine whether the properties are represented without supposing such structures by targeting semantic properties directly rather than in relation to other concepts. 

Several approaches have attempted to derive properties collected in property norm datasets from the distribution in naturally occurring texts \citep{Kelly2014, Baroni2010a,Barbu2008}. Whereas these approaches yield indications about the potential of distributional models, they do not go beyond full-vector proximity on a low-dimensional SVD model or context words in a transparent, high-dimensional count model. Their focus lies on detecting informative contexts. We follow the idea behind this approach and make a human-elicited property dataset that is created in the same tradition, but larger. Our approach goes beyond the previous work in two ways: first, we add gold negative examples which allows us to go beyond testing for salient properties. Second, we compare full vector proximity to the outcome of a classifier which allows us to verify whether the property is captured for entities that share the property, but are not similar otherwise.

A few other studies go beyond full vector comparisons, moving towards the interpretation of word embedding dimensions. \citet{Tsvetkov2015,Tsvetkov2016} evaluate word embeddings by measuring the correlation between word embedding vectors and count vectors representing co-occurrences of words with WordNet supersenses. While they show that their results have a higher correlation with results obtained from extrinsic evaluations than standardly used intrinsic evaluations, they do not provide insights into what kind of semantic information is represented well. \citet{Yaghoobzadeh2016} decompose distributional vectors into individual linguistic aspects by means of a supervised classification approach to test which linguistic phenomena are captured by embeddings. They test their approach on an artificially created corpus and do not provide insights into specific semantic knowledge. \citet{Faruqui2015} transform learned embedding matrices into sparse matrices to make them more interpretable, which is complementary to our approach.  

Previous studies provide (indicative) support for the hypothesis that embeddings lack information people get from other modalities than language. \citet{Fagarasan2015} present a method to ground embedding models in perceptual information by mapping distributional spaces to semantic spaces consisting of feature norms. Several approaches to boosting distributional models with visual information show that the additional information improves the performance of word embedding vectors \citep{Roller2013,Lazaridou2014}. Whereas this indicates that word embedding models lack visual information, it does not show to what extent different types of properties are encoded. 
The method proposed in this paper is, to the best of our knowledge, the first approach specifically designed to identify what semantic knowledge is captured in word embeddings. We are not aware of earlier work that provides explicit hypotheses about the kind of information we expect to learn from distributional vectors, making this the first attempt to confirm these hypotheses experimentally.


\section{Method}\label{sec:method}

The core of our evaluation consists of testing whether nearest neighbors and classifiers are capable of identifying which embeddings encode a given semantic property. We first describe the dataset and then present the procedure we apply. We complete this section with our hypotheses about the outcome of our evaluation.

\subsection{Extended CSLB Data} \label{subsec.data}

We use the Centre for Speech, Language and the Brain concept property norms dataset \cite[henceforth CSLB]{Devereux2014}. This dataset follows the tradition of the sets created by \newcite{McRae2005, Vinson2008} and used in \newcite{Kelly2014, Baroni2010a,Barbu2008} and is the largest available semantic property dataset we are aware of. In the collection process, human subjects were given concrete and mostly monosemous concepts and asked to provide a set of semantic features. Polysemous concepts were disambiguated. Properties were elicited by cues such as \textit{has}, \textit{is}, \textit{does} and \textit{made\_of}. An empty slot was provided to fill in other relations. The dataset comprises 638 annotated concepts, each of which was presented to 30 participants. Properties listed by at least two participants are included in the published set. 

We select features associated with at least 20 concepts. In an exploratory experiment, we count all concepts for which the target feature is listed as positive examples and all other concepts as negative examples. However, the fact that people did not list a property does not necessarily mean that a given concept is a negative example of it. For instance: \textit{falcon} is described by \textit{is\_a\_bird}, but not by \textit{is\_an\_animal}.

For proper evaluation, the CSLB dataset should be extended with verified negative examples. We apply two methods to add both positive and (verified) negative properties to CSLB. First, we select properties that necessarily imply the target property (e.g.\ \textit{is\_a\_bird} implies \textit{is\_an\_animal}) or necessarily exclude the target property (e.g.\ \textit{is\_food} almost certainly excludes \textit{has\_wheels}). We both manually inspect the extended sets of positive and negative examples per selected property to exclude remaining noise independently, resolving disagreements after discussion.\footnote{All annotations, guiding principles as well as notes about resolving discussions can be found at \url{https://cltl.github.io/semantic_space_navigation}.} 

The resulting dataset has the disadvantage that negative examples largely consist of the same specific categories, e.g.\ negative examples of \textit{has\_wheels} are food, animals and plants. Based on these examples, we cannot tell whether the classifier performs well because embeddings encode the property of having wheels or because it can distinguish vehicles from food, animals and plants. We therefore need to expand the dataset so that it includes diverse negative and positive examples and preferably positive and negative examples that are closely related in semantic space. 

Ultimately, we want to verify and increase the entire dataset and distinguish between things that always or typically have a property (e.g.\ $bike$ $has\_wheels$, $banana$ $is\_yellow$), things that can have a property (e.g.\ $bikini$ - $is\_pink$, $plate$ - $made\_of\_metal$) and things that do normally not have a property (e.g.\ $grape$ - $does\_kill$, $beer$ - $is\_pink$).
We set up a crowdsourcing task in which we ask participants whether a property applies to a word. Possible answers are \textit{yes}, \textit{mostly}, \textit{possibly} and \textit{no}. 

This crowdsourcing method has currently been applied to a selection of property-concept pairs that were labeled as false-positives by at least one of our approaches in the initial setup. In addition, we extend the property-concept pairs given to crowd workers by collecting the nearest neighbors of the property centroid and a number of seed words. We aim at (1) identifying negative examples that have a high cosine similarity to positive examples in the dataset and (2) including a broader variety of words. This nearest-neighbors strategy explicitly aims at collecting words that are highly similar to positive examples of a property but are not associated with it. For instance, in order to extend the concept set for the property \textit{has\_wheels}, we used the seed words \textit{car}, \textit{sledge}, and \textit{ship}.\footnote{The details about our selection and full lists of seed words are provided with our code (see link in Footnote 1).}

In the experiments reported in this paper, we only consider properties that clearly apply to a concept as positive examples (the \textit{yes} and \textit{mostly} cases) and properties that clearly do not apply as negative examples, leaving disputable cases and the cases that \textit{possibly} apply for future work. We manually checked cases of disagreement in the crowd data and selected or removed data based on these criteria.\footnote{Some difference in judgment are clearly the result of lack of knowledge (e.g.\ not knowing a something is an animal). The original outcome of the crowd and final resulting test are provided on the github repository associated with this paper.}

\subsection{Classification approaches} \label{subsec.class}

We use the pretrained Word2vec model based on the Google News corpus.\footnote{\url{https://code.google.com/archive/p/word2vec/}} The underlying architecture is a skip-gram with negative sampling model \citep{Mikolov2013}, which learns word vectors by predicting the context given a word. 

The overall goal is to investigate whether word vectors capture specific semantic properties or not. 
We start from the assumption that classifiers can learn properties that are represented in the embedding in a binary classification task. We apply supervised classification to see whether a logistic regression classifier or a neural network are capable of distinguishing embeddings of words that have a specific semantic property from those which do not. Specifically, we use embedding vectors corresponding to words associated or not associated with a semantic target-property (i.e.\ positive and negative examples) as input for a binary classifier and test whether the classifier can learn to distinguish embeddings of words that have the property from those who do not. However, word embeddings also capture semantic similarity. If a property is shared by similar entities (e.g.\ most animals with a beak are birds), the classifiers may perform well because of this similarity rather than identifying the actual property. We therefore compare the performance of classifiers to the performance of an approach based on full vector similarity. If only the classifiers score well, this provides an indication that the embedding captures the property. If both methods perform poorly this could mean that the property is not captured.\footnote{Given the size and balance of our dataset as well as the lack of fine-tuning, we remain careful not to draw firm conclusions at this point.}

\subsubsection*{Supervised classification}

As the datasets are limited in size, we evaluate by applying a leave-one-out approach. We employ two different supervised classifiers, which we expect to differ in performance. As a `vanilla' approach, we use a logistic regression classifier with default settings as implemented in SKlearn. This type of classifier is also used in \newcite{Drozd2016} to detect words of similar categories in an improved analogy model. 

In addition, we use a basic neural network. Meaningful properties may not always be encoded in individual patterns, but rather arise from a combination of activated dimensions. This is not captured well by a logistic regression model, as it can only react to individual dimensions. In contrast, the neural network can learn from patterns of dimensions. We use a simple multi-layer perceptron (as implemented in SKlearn\footnote{\url{http://scikit-learn.org/stable/index.html}}) with a single hidden layer. We calculate the number of nodes in the hidden layer as follows: (number of input dimensions + number of output dimensions) * 1/3. The pretrained Google News vectors have 300 dimensions, resulting in a hidden layer of 100 nodes. We use the recommended settings for small datasets. No parameter tuning was conducted so far due to the limited size of the datasets and the use of a leave-one-out evaluation strategy. We present the runs of several models, as the neural network can react to the order in which the examples are presented as well as the randomly assigned vectors for initialization. While the performance of the model could be optimized further by experimenting with the settings, we find that the set-up presented here already outperforms the logistic regression classifier in many cases. 

\subsubsection*{Full vector similarity}

To show that supervised classification can go beyond full vector comparison in terms of cosine similarity, we compare the performance of the classifiers to an n-nearest neighbors approach. We calculate the centroid vector of all positive examples in the training set. The training set consists of all positive examples in the leave-one-out split except for the one we are testing on. We then consider its n-nearest neighbors measured by their cosine distance to the centroid as positive examples. We vary n between 100 and 1,000 in steps of 100. We report the performance of the optimal number of neighbors for each property (which varies per property). In future work, we will add more fine-grained steps and investigate the performance of a classifier using the cosine similarity of words to the centroid as a sole feature. 

\subsubsection*{Variety approximation}

The performance of the approaches outlined above depends on to the variety of words associated with a property. We approximate this variety by calculating the average cosine similarity of words associated with a property to one-another. This is done by averaging over the cosine similarities between all possible pairs of words. A high average cosine similarity means that the words associated with a concepts tend to be close to each other in the space, which should mostly apply to words associated with taxonomic categories. In contrast, a low average cosine means a high diversity, which should largely apply to general descriptions.

\subsection{Specific hypotheses}
\label{subsec.hypotheses}

We select a number of properties for closer investigation based on the clean and extended dataset described in Section~\ref{subsec.data}. We first formulated the hypotheses independently, before discussing and specifying them.\footnote{Details can be found on the paper's github repository.} Table~\ref{tab.hyp} summarizes the agreed upon expectations. The hypotheses can be categorized in the following way:

\subsubsection*{Sparse Textual Evidence}

We select properties of which we expect that textual evidence is too sparse to be represented by distributional vectors. The properties \textit{is\_black}, \textit{is\_yellow}, \textit{is\_red} and \textit{made\_of\_wood} have little impact on the way most entities belonging to that class interact with the world. We expect that the only textual evidence indicating them are individual words denoting the properties themselves (e.g.\ \textit{red}, \textit{black}, \textit{wooden})\footnote{In the case of \textit{made\_of\_wood}, the evidence may be a bit broader, as it might be indicated by different types of wood occurring in the context of furniture.} and it is unclear how often they are mentioned explicitly. It may, however, be the case that certain subcategories in the datasets are learned regardless of this sparsity, because they happen to coincide with more relevant taxonomic categories such as \textit{red fruits}. 

\subsubsection*{Fine-grained Distinctions in Larger Categories}

We expect that a supervised classifier may be able to make more fine-grained distinctions between examples of the same category when these differences are relevant for the way they interact with the world. We select two properties that introduce crucial distinctions in larger categories: \textit{has\_wheels} and \textit{is\_found\_in\_seas}. The former applies to a sub-group of vehicles and may be apparent in certain behaviors and contexts only applying to these vehicles (rolling, street, etc). The latter applies to animals, plants and other entities found in water, but it is unclear whether textual evidence is enough to distinguish between seawater and fresh water. 

\subsubsection*{Mixed Groups}

We expect that a supervised machine learning approach can find positive examples of a property that are not part of the most common class in the training set. For instance, the majority of positive examples for \textit{is\_dangerous} and \textit{does\_kill} refer to weapons or dangerous animals. We expect the classifier to (1) find positive examples from less well represented groups and (2) be able to distinguish between positive and negative examples of a well-represented category (e.g.\ \textit{rhino} v.s.\ \textit{hippo} for killing). For the property \textit{is\_used\_in\_cooking}, the example words refer to food items as well as utensils. We expect that classifiers can distinguish between cooking-related utensils and other tools.

\begin{table}
\begin{center}
\small{
\begin{tabular}{l  r   }
\cmidrule[\heavyrulewidth](lr){1-2}
Property & learnable property \\
\cmidrule(lr){1-2}
is\_an\_animal & yes   \\
is\_food & yes \\
is\_dangerous & yes \\
does\_kill & yes \\
is\_used\_in\_cooking & yes \\
has\_wheels & possibly \\
is\_found\_in\_seas & possibly \\
is\_black & no \\
is\_red & no \\
is\_yellow & no \\ 
made\_of\_wood & no \\ 
\cmidrule[\heavyrulewidth](lr){1-2}
\end{tabular}
\caption{Hypotheses about whether selected semantic properties can be learned by a supervised classifier }\label{tab.hyp}}
\end{center}
\vspace{-0.6cm}
\end{table}

\subsubsection*{Polysemy}

We expect that machine learning can recognize vector dimensions indicating properties applying to different senses of a word, whereas the nearest-neighbors approach simply assigns the word to its dominant class. For instance, we expect that word vectors that can be used to describe animals as well as food (e.g.\ \textit{chicken}, \textit{rabbit} or \textit{turkey}) record evidence of both contexts, but end up closer to one of the categories. A supervised machine learning approach should be able to find the relevant dimensions regardless of the cosine similarity to one of the groups and classify the word correctly. We test this by training on a set of monosemous words (animals and food items) and test on a set of polysemous and monosemous examples.

\section{Experimental set-up and results}\label{sec:experiments}

\subsection{Concept diversity vs performance}

We first investigate the relation between performance and diversity of concepts associated with a property on the full, noisy dataset using a leave-one-out approach. 
Table~\ref{tab.res-naive} shows a selection of the f1-scores achieved on properties in the CSLB dataset in relation to the average cosine similarity of the associated words. A high average cosine similarity means that the concepts overall have similar vector representations and can thus be seen as having a low diversity. The results of the Spearman Rank correlation clearly indicate that scores achieved by nearest neighbors correlate more strongly with the average cosine than the two supervised classification approaches. 

\begin{table}
\centering
\small{
\begin{tabular}{p{2cm} rrrrl}
\toprule
feature                           & cos  & f1-neigh & f1-lr & f1-net & type \\
\midrule
is\_heavy                         & 0.15 & 0.15                  & 0.17                    & 0.21                           & op   \\
is\_strong                        & 0.15 & 0.13                  & 0.13                    & 0.34                           & e    \\
is\_thin                          & 0.16 & 0                     & 0.05                    & 0.1                            & vp   \\
is\_hard                          & 0.16 & 0.15                  & 0.08                    & 0.26                           & op   \\
is\_expensive                     & 0.16 & 0                     & 0.28                    & 0.37                           & e    \\
... & ... & ... & ...  & ... \\
is\_black                         & 0.2  & 0.29                  & 0.23                    & 0.24                           & vp   \\
is\_electric                      & 0.21 & 0.48                  & 0.5                     & 0.69                           & vp   \\
is\_dangerous                     & 0.21 & 0.53                  & 0.57                    & 0.59                           & e    \\
is\_colourful                     & 0.21 & 0.14                  & 0.25                    & 0.32                           & vp   \\
is\_brown                         & 0.21 & 0.13                  & 0.22                    & 0.33                           & vp   \\
has\_a\_handle \_handles           & 0.22 & 0.44                  & 0.57                    & 0.58                           & p    \\
has\_a\_seat \_seats               & 0.22 & 0.43                  & 0.3                     & 0.48                           & p    \\
does\_smell \_is\_smelly           & 0.22 & 0.08                  & 0.15                    & 0.37                           & op   \\
made\_of\_glass                   & 0.22 & 0.29                  & 0                       & 0.28                           & vp   \\
has\_a\_point                     & 0.23 & 0.38                  & 0.23                    & 0.47                           & p    \\
does\_protect                     & 0.24 & 0.38                  & 0.26                    & 0.37                           & f    \\
is\_yellow                        & 0.24 & 0.22                  & 0                       & 0.23                           & vp   \\
is\_soft                          & 0.24 & 0.12                  & 0                       & 0.16                           & op   \\
is\_red                           & 0.25 & 0.34                  & 0.13                    & 0.27                           & vp   \\
is\_fast                          & 0.25 & 0.3                   & 0.31                    & 0.48                           & vp   \\
is\_tall                          & 0.25 & 0.43                  & 0.57                    & 0.65                           & vp   \\
is\_a\_tool                       & 0.26 & 0.5                   & 0.51                    & 0.47                           & t    \\
... & ... & ... & ...  & ... \\
is\_a\_weapon                     & 0.3  & 0.74                  & 0.56                    & 0.63                           & t    \\
is\_green                         & 0.31 & 0.45                  & 0.45                    & 0.45                           & vp   \\
has\_a\_ blade\_blades             & 0.32 & 0.68                  & 0.65                    & 0.74                           & p    \\
is\_worn                          & 0.32 & 0.47                  & 0.86                    & 0.9                            & f    \\
has\_wheels                       & 0.32 & 0.82                  & 0.83                    & 0.87                           & p    \\
is\_found \_in\_kitchens           & 0.33 & 0.56                  & 0.73                    & 0.76                           & e    \\
does\_fly                         & 0.33 & 0.57                  & 0.76                    & 0.76                           & f    \\
has\_a\_tail                      & 0.33 & 0.53                  & 0.68                    & 0.69                           & p    \\
is\_an\_animal                    & 0.33 & 0.64                  & 0.76                    & 0.78                           & t    \\
is\_eaten\_edible                 & 0.33 & 0.37                  & 0.88                    & 0.85                           & f    \\
has\_four\_legs                   & 0.34 & 0.67                  & 0.66                    & 0.66                           & p    \\
is\_a\_vehicle                    & 0.34 & 0.76                  & 0.69                    & 0.79                           & t    \\
does\_eat                         & 0.34 & 0.68                  & 0.71                    & 0.68                           & f    \\
... & ... & ... & ...  & ... \\
has\_a\_beak                      & 0.37 & 0.63                  & 0.83                    & 0.87                           & p    \\
made\_of\_cotton                  & 0.37 & 0.68                  & 0.56                    & 0.64                           & vp   \\
has\_roots                        & 0.37 & 0.3                   & 0.65                    & 0.72                           & p    \\
is\_a\_mammal                     & 0.37 & 0.69                  & 0.85                    & 0.86                           & t    \\
does\_grow                        & 0.37 & 0.52                  & 0.81                    & 0.81                           & e    \\
is\_a\_plant                      & 0.37 & 0.43                  & 0.63                    & 0.64                           & t    \\
has\_leaves                       & 0.37 & 0.41                  & 0.71                    & 0.78                           & p    \\
... & ... & ... & ...  & ... \\
has\_pips\_seeds                  & 0.47 & 0.5                   & 0.08                    & 0.46                           & p    \\
is\_juicy                         & 0.5  & 0.71                  & 0.48                    & 0.56                           & op   \\
is\_a\_vegetable                  & 0.52 & 0.78                  & 0.75                    & 0.81                           & t    \\
is\_played \_does\_play            & 0.53 & 0.9                   & 0.98                    & 0.98                           & f    \\
does\_make\_music                 & 0.55 & 0.89                  & 0.95                    & 0.92                           & f    \\
spearman-r                        &      & 0.72                  & 0.52                    & 0.59                           &  \\  
\bottomrule

\end{tabular}
\caption{Performance of different approaches in relation to the average cosine similarity of words associated with a property (cos). The last row shows the Spearman Rank correlation between f1-scores and average cosine similarity. Property types are listed under type (p = part, vp = visual-perceptual, op = other-perceptual, e = encyclopaedic, f = functional, t = taxonomic). }\label{tab.res-naive}}
\vspace{-0.5cm}
\end{table}

\subsection{Outcome Specific Hypotheses}

We carry out further experiments on a small extended and clean subset, consisting of carefully selected negative examples from the CSLB dataset and crowd annotations validated by the authors. The distribution of positive and negative examples per property is shown in Table~\ref{tab.distribution}. For some properties, the sets derived from the CSLB norms alone have an imbalanced distribution of negative examples over semantic categories, as they were selected by means of logical exclusion (e.g. concepts listed under \textit{has\_wheels} have been selected as negative examples of \textit{is\_food}). Therefore, we add the more balanced but smaller datasets created by crowd-judgments only where enough judgments have been collected. We created additional sets for words part of the food-animal polysemy to test whether supervised classifiers can successfully predict semantic properties of various senses of polysemous words. In the following sections, we will outline the most striking results. Most results confirm, but some contradict our initial hypotheses. 

\begin{table}
\centering
\small{
\begin{tabular}{lrr}
\toprule
Property & pos & neg \\
\midrule
full\_does\_kill            & 101 & 69  \\
crowd\_does\_kill            & 67       & 49 \\
full\_has\_wheels           & 79  & 349 \\

full\_is\_black             & 42  & 89  \\

full\_is\_dangerous         & 177 & 104 \\
crowd\_is\_dangerous         & 131      & 84 \\

full\_is\_found\_in\_seas   & 83  & 72  \\
crowd\_is\_found\_in\_seas   & 47       & 28  \\

full\_is\_red               & 29  & 80  \\

full\_is\_used\_in\_cooking & 142 & 61  \\

full\_is\_yellow            & 24  & 68  \\

full\_made\_of\_wood        & 87  & 282 \\

full\_is\_an\_animal\_test  & 37  & 20  \\
full\_is\_an\_animal\_train & 166 & 77  \\
full\_is\_food\_test        & 37  & 20  \\
full\_is\_food\_train       & 97  & 146 \\ 
\bottomrule
\end{tabular}
\caption{Class distribution in dataset consisting of the clean datasets derived from the CSLB set and the additional crowd judgments (marked \textit{full\_}). For some properties, we included the dataset consisting of crowd-judgments only, as it is more balanced across semantic categories than the full set (marked \textit{crowd\_}). For all properties, a leave-one-out approach was applied to evaluation except for \textit{is\_animal} and \textit{is\_food}.}\label{tab.distribution}}
\end{table}

Table~\ref{tab.res1} shows the f1-scores on the full clean datasets. As hypothesized, the color properties \textit{is\_yellow} and \textit{is\_red} perform low in all approaches, with slightly better results yielded by supervised learning.

The properties involved in functions and activities or with high impact on the interaction of entities with the world all perform highly in the classification approaches. For \textit{does\_kill}, \textit{is\_dangerous} and \textit{is\_used\_in\_cooking}, there is a large difference between the best nearest neighbors approach and the best classification approach (between 60 and 19 points), indicating that the classification approaches are able to infer more information from individual dimensions than is provided by full vector similarity. The property \textit{is\_dangerous} has, as can be expected, a particularly high diversity of associated words (comparable to the colors). \textit{Has\_wheels} and \textit{is\_found\_in\_seas} can be expected to have high correlations with other taxonomic categories (fish and water animals, vehicles), which is reflected in the lower diversity and comparatively high nearest neighbor performance. 

Cases contradicting our expectations are the visual properties \textit{is\_black} and \textit{made\_of\_wood}. Both have comparatively high classification performance with a big difference to the nearest neighbor results. Most likely, this is due to a category bias in the negative examples. For instance, a large portion of the negative examples for \textit{is\_made\_of\_wood} consist of animals and food. In the dataset for \textit{is\_black}, a large proportion of the positive examples consists of animals. A classifier can perform highly by simply learning to distinguish these two categories from the rest. 

The biases in semantic classes mentioned above partially result from the way we generated the negative examples from the original CSLB dataset. This means that a classifier may learn to distinguish two semantic categories rather than being able to find vector dimensions indicative of the target property. We therefore also present selected results on crowd-only datasets shown in Table~\ref{tab.res1}, which do not have this bias. It can be observed that for all three properties,\footnote{We only included properties for which we had enough positive and negative examples in our set} the performance of the classification approaches drops marginally, whereas it rises for nearest neighbors. 

We investigate the outcome on a number of individual examples to gain more insights into whether the subtle differences hypothesized in Section~\ref{sec:method} hold. Since we only formulate a general hypothesis for Sparse Textual Evidence, we do not dive deeper into the results for that category here.

\subsubsection*{Fine-Grained Category Distinctions}

The full clean \textit{has\_wheels} dataset includes a number of instances for which the classifiers can make more fine-grained distinctions than nearest neighbors. As hypothesized, classifiers, in contrast to nearest-neighbors, can recognize that neither \textit{sled} nor a \textit{skidoo} have wheels, but a \textit{unicycle} a \textit{limousine}, a \textit{train}, \textit{carriage}, an \textit{ambulance}, a \textit{porsche} do. Another fine-grained distinction can be identified in the \textit{is\_found\_in\_seas} crowd-only set: \textit{Sculpin} is correctly identified as a seawater fish by all classifiers but not by nearest-neighbors.

\subsubsection*{Mixed Groups}

Whereas nearest neighbors predominantly identify weapons as \textit{is\_dangerous} in the crowd-only set, the classifiers go beyond this category. The neural network approach correctly identifies that \textit{imitation pistol}, \textit{imitation handgun}, and \textit{screwdriver} are negative examples of \textit{is\_dangerous}. Furthermore, no animals are labeled as dangerous based on proximity to the centroid, but the classifiers are able to distinguish between some dangerous and non-dangerous animals (e.g.\ \textit{rhinoceros} is labeled positive, while \textit{giraffe} and \textit{zebra} are labeled as negative). All three classifiers recognize that \textit{meth}, \textit{cocaine} and \textit{oxycodone} are considered dangerous substances, despite the fact that they are far away from the centroid of dangerous things. Of the only two disease-like concepts, \textit{Hepatitis C} and \textit{allergy}, the former is recognized by all classifiers and the latter only by logistic regression. The performance on the smaller, but also weapon-dominated \textit{does\_kill} crowd-only set is comparable, but the variety of atypical cases is lower. Among the only two disease-related items, \textit{dengue} is identified by all classifiers and \textit{dengue virus} only by the neural network. 

In the crowd-only \textit{is\_found\_in\_seas} set, \textit{seabird} and \textit{gannet} are correctly labeled as positive, even though  positive examples almost exclusively consist of fish or underwater-animals, whereas the negative examples encompass a vast variety of animals, including \textit{bird} and some freshwater fish. 

\subsubsection*{Polysemy}

For polysemy between food and animals (Table~\ref{tab.res1}), we observe that when trained on pure animal and food words and tested on polysemous animal and food words, the classifiers perform highly with a large difference to nearest neighbors. For food versus pure animal words, the classifier performance is much lower. We expect the extremely low nearest neighbor performance to be due to the fact that the centroid is calculated over pure food items (without a single animal-related item, not even culinary meat terms such as \textit{pork} or \textit{beef}) which is far away from the animal-region in the space. Despite the classifiers outperforming nearest neighbors, the outcome does not confirm our original hypotheses. We expected that the classifiers could identify that edible animals have both animal properties and food properties, but upon inspection of the results, the classifiers only identified entities with a predominant animal sense correctly as animals and those with a predominant food sense correctly as food.

\begin{table}
\centering
\small{
\begin{tabular}{p{2cm}rrrrr}
\toprule
property  &  av-cos &  neigh &  lr &  net1 &  net2  \\     
\midrule
full\_is\_yellow          &    0.23 &    0.19 &  0.47 &  0.64 &  0.64\\
full\_is\_used\_in \_cooking &    0.37 & 0.29  &  0.98 &  0.98 &  0.98 \\
full\_is\_black           &    0.19 & 0.35 & 0.75 &  0.77 &  0.77  \\
full\_is\_red             &    0.23 & 0.36 &  0.51 &  0.54 &  0.52   \\
full\_is\_dangerous       &    0.24 & 0.58 &  0.88 &  0.88 &  0.87   \\
crowd\_is\_dangerous       &    0.26 & 0.61  & 0.86 &  0.86 &  0.86   \\
full\_has\_wheels         &    0.38 &  0.90 &  0.96 &  0.96 &  0.95   \\
full\_is\_found\_in\_seas   &    0.44 & 0.87 &  0.97 &  0.98 &  0.98   \\
crowd\_is\_found \_in\_seas   &    0.50 & 0.87 &  0.94 &  0.96 &  0.96  \\
full\_does\_kill          &    0.27 & 0.67 & 0.83 &  0.86 &  0.82    \\
crowd\_does\_kill          &    0.30 & 0.70 &  0.82 &  0.84 &  0.80   \\
full\_made\_of\_wood       &    0.17 & 0.14 &  0.84 &  0.85 &  0.85    \\
full\_is\_food\_test      &    0.37 & 0.00 &  0.36 &  0.36 &  0.36   \\
full\_is\_an \_animal\_test &    0.37 & 0.52 &  0.88 &  0.88 &  0.88  \\
\bottomrule
\end{tabular}
\caption{F1 scores achieved by logistic regression (lr) two runs of a neural net classifier (net1 and net2 and the n-best nearest neighbors evaluated with leave-one-out on the full datasets (marked as \textit{full\_} and the crow-only sets (marked as \textit{crowd\_}). }\label{tab.res1}.} 
\end{table}

\section{Discussion \& Future Work}\label{sec:discussion}

The experiments presented in this approach have several limitations. First, our semantic datasets are still limited in size. Second, the implication method we applied to generate negative examples led to biases for some properties where most negative examples belong to a small set of (taxonomic) classes. Third, no parameter tuning has been carried out so far. Careful parameter tuning would ensure that the best possible classification approaches are chosen and that the obtained results truly exploit the informative power of the embeddings. Due to the limited size of the dataset and the leave-one-out approach to evaluation, this has not been possible in this preliminary study. Fourth, the experiments presented here only concern a small subsection of semantic properties too limited to draw general conclusions. 

Despite these limitations, our results provide preliminary insights that lead us to conclude that the overall idea behind our methods works and opens up promising directions for future work. We first aim to address the limitations of the current dataset. We intend to incorporate other sets designed for similar insights, such as the analogies presented in \citet{Drozd2016} and the SemEval 2018 discriminative property set \citep{krebs2018}. In addition, we plan to extend and refine the sets with crowd annotations asking for graded judgments (e.g.\ a property can \textit{mostly} or \textit{possibly} apply) and exploit these judgments in future experiments.

Once we created a bigger and more balanced dataset, we can carry out experiments on different train and test splits in order to overcome the limitations of the leave-one-out evaluation. Furthermore, we will apply careful parameter tuning on a development set in order to ensure our results are representative of the information captured by the embeddings. The increased size of the set will allow us to conduct more experiments that take the distributions of semantic categories in the splits into account as was done for the polysemy set. This way, we ensure that we do not train on examples that belong to the same semantic category as the ones in the test set.

Going beyond the method introduced in this paper, we plan on investigating the type of information encoded in linguistic context by testing which properties can be learned from textual context directly. In addition, applying the method presented by \citet{Faruqui2015} may provide stronger indications about the information represented by word embedding dimensions. Adding these to experiments allows us to trace which information is provided by the context and what ends up being present in word embeddings. 

\section{Conclusion}\label{sec:conclusion}

The main contribution of this paper is that it introduces a new method aimed at investigating the kind of semantic information captured by word embedding vectors. We have taken the first steps towards constructing a dataset suitable for this investigation on the basis of an existing dataset of human-elicited semantic properties. We introduced a set of hypotheses concerning which semantic properties are captured by embeddings and presented exploratory experiments verifying them.

The current results are limited by the size and balance of our dataset, as discussed in detail in the previous section. Nevertheless, we can report preliminary insights based on our experiments. We show that classifiers, in particular neural networks, can identify which entities have a specific property in cases where this does not follow from general similarity or the overall semantic class the entity belongs to. This can be seen as a first indication that (some) semantic properties are encoded in individual (patterns of) vector dimensions, which can be identified. 

The results on the extended datasets partly confirm that visual properties are not well represented by embeddings, while properties relating to function (e.g.\ cooking, having wheels) and interactions with other entities (e.g.\ being dangerous or killing) tend to be represented well. Some of these indications could be the result of the bias in our current dataset, but others have been confirmed on the smaller crowd-only sets for properties with enough available data (\textit{is\_dangerous} and \textit{does\_kill}). Further evidence is provided by the full dataset for \textit{has\_wheels} which encompasses a large group of vehicles to which the property does not apply. In addition, we support these indications by qualitative insights through examples of the kinds of distinctions made by the classifiers, but not the nearest neighbor approach. Results achieved for polysemous words and two visual properties currently do not confirm our hypotheses.

\section*{Acknowledgements}

This research is funded by the PhD in the Humanities Grant provided by the Netherlands Organization of Scientific Research (Nederlandse Organisatie voor Wetenschappelijk Onderzoek, NWO) PGW.17.041 awarded to Pia Sommerauer and NWO VENI grant 275-89-029 awarded to Antske Fokkens. We would like to thank anonymous reviewers for feedback and Piek Vossen for feedback and discussion that helped improve this paper. All remaining errors are our own.

\bibliography{emnlp}

\begin{thebibliography}{25}
\expandafter\ifx\csname natexlab\endcsname\relax\def\natexlab#1{#1}\fi

\bibitem[{Barbu(2008)}]{Barbu2008}
Eduard Barbu. 2008.
\newblock Combining methods to learn feature-norm-like concept descriptions.
\newblock In \emph{Proceedings of the ESSLLI Workshop on Distributional Lexical
  Semantics}, pages 9--16.

\bibitem[{Baroni et~al.(2010)Baroni, Murphy, Barbu, and Poesio}]{Baroni2010a}
Marco Baroni, Brian Murphy, Eduard Barbu, and Massimo Poesio. 2010.
\newblock Strudel: A corpus-based semantic model based on properties and types.
\newblock \emph{Cognitive science}, 34(2):222--254.

\bibitem[{Devereux et~al.(2014)Devereux, Tyler, Geertzen, and
  Randall}]{Devereux2014}
Barry~J. Devereux, Lorraine~K. Tyler, Jeroen Geertzen, and Billi Randall. 2014.
\newblock The centre for speech, language and the brain (cslb) concept property
  norms.
\newblock \emph{Behavior research methods}, 46(4):1119--1127.

\bibitem[{Drozd et~al.(2016)Drozd, Gladkova, and Matsuoka}]{Drozd2016}
Aleksandr Drozd, Anna Gladkova, and Satoshi Matsuoka. 2016.
\newblock Word embeddings, analogies, and machine learning: Beyond king-man+
  woman= queen.
\newblock In \emph{COLING}, pages 3519--3530.

\bibitem[{Fagarasan et~al.(2015)Fagarasan, Vecchi, and Clark}]{Fagarasan2015}
Luana Fagarasan, Eva~Maria Vecchi, and Stephen Clark. 2015.
\newblock From distributional semantics to feature norms: grounding semantic
  models in human perceptual data.
\newblock In \emph{Proceedings of the 11th International Conference on
  Computational Semantics}, pages 52--57.

\bibitem[{Faruqui et~al.(2015)Faruqui, Tsvetkov, Yogatama, Dyer, and
  Smith}]{Faruqui2015}
Manaal Faruqui, Yulia Tsvetkov, Dani Yogatama, Chris Dyer, and Noah~A. Smith.
  2015.
\newblock Sparse overcomplete word vector representations.
\newblock In \emph{Proceedings of ACL}.

\bibitem[{Gladkova and Drozd(2016)}]{Gladkova2016a}
Anna Gladkova and Aleksandr Drozd. 2016.
\newblock Intrinsic evaluations of word embeddings: What can we do better?
\newblock In \emph{Proceedings of the 1st Workshop on Evaluating Vector-Space
  Representations for NLP}, pages 36--42.

\bibitem[{Gladkova et~al.(2016)Gladkova, Drozd, and Matsuoka}]{Gladkova2016b}
Anna Gladkova, Aleksandr Drozd, and Satoshi Matsuoka. 2016.
\newblock Analogy-based detection of morphological and semantic relations with
  word embeddings: what works and what doesn't.
\newblock In \emph{Proceedings of the NAACL Student Research Workshop}, pages
  8--15.

\bibitem[{Kelly et~al.(2014)Kelly, Devereux, and Korhonen}]{Kelly2014}
Colin Kelly, Barry Devereux, and Anna Korhonen. 2014.
\newblock Automatic extraction of property norm-like data from large text
  corpora.
\newblock \emph{Cognitive Science}, 38(4):638--682.

\bibitem[{Krebs et~al.(2018)Krebs, Lenci, and Paperno}]{krebs2018}
Alicia Krebs, Alessandro Lenci, and Denis Paperno. 2018.
\newblock Semeval-2018 task 10: Capturing discriminative attributes.
\newblock In \emph{Proceedings of The 12th International Workshop on Semantic
  Evaluation}, pages 732--740.

\bibitem[{Lazaridou et~al.(2014)Lazaridou, Bruni, and Baroni}]{Lazaridou2014}
Angeliki Lazaridou, Elia Bruni, and Marco Baroni. 2014.
\newblock Is this a wampimuk? cross-modal mapping between distributional
  semantics and the visual world.
\newblock In \emph{Proceedings of the 52nd Annual Meeting of the Association
  for Computational Linguistics (Volume 1: Long Papers)}, volume~1, pages
  1403--1414.

\bibitem[{Lazaridou et~al.(2013)Lazaridou, Vecchi, and Baroni}]{Lazaridou:2013}
Angeliki Lazaridou, Eva~Maria Vecchi, and Marco Baroni. 2013.
\newblock Fish transporters and miracle homes: How compositional distributional
  semantics can help np parsing.
\newblock In \emph{Proceedings of the 2013 Conference on Empirical Methods in
  Natural Language Processing}, pages 1908--1913.

\bibitem[{Levy and Goldberg(2014)}]{Levy2014}
Omer Levy and Yoav Goldberg. 2014.
\newblock Linguistic regularities in sparse and explicit word representations.
\newblock In \emph{Proceedings of the eighteenth conference on computational
  natural language learning}, pages 171--180.

\bibitem[{Linzen(2016)}]{Linzen2016}
Tal Linzen. 2016.
\newblock Issues in evaluating semantic spaces using word analogies.
\newblock In \emph{Proceedings of the 1st Workshop on Evaluating Vector-Space
  Representations for NLP}, pages 13--18.

\bibitem[{McRae et~al.(2005)McRae, Cree, Seidenberg, and McNorgan}]{McRae2005}
Ken McRae, George~S Cree, Mark~S Seidenberg, and Chris McNorgan. 2005.
\newblock Semantic feature production norms for a large set of living and
  nonliving things.
\newblock \emph{Behavior research methods}, 37(4):547--559.

\bibitem[{Mikolov et~al.(2013)Mikolov, Yih, and Zweig}]{Mikolov2013}
Tomas Mikolov, Wen-tau Yih, and Geoffrey Zweig. 2013.
\newblock Linguistic regularities in continuous space word representations.
\newblock In \emph{HLT-NAACL}, volume~13, pages 746--751.

\bibitem[{Rogers et~al.(2017)Rogers, Drozd, and Li}]{Rogers2017}
Anna Rogers, Aleksandr Drozd, and Bofang Li. 2017.
\newblock The (too many) problems of analogical reasoning with word vectors.
\newblock In \emph{Proceedings of the 6th Joint Conference on Lexical and
  Computational Semantics (* SEM 2017)}, pages 135--148.

\bibitem[{Roller and Schulte~im Walde(2013)}]{Roller2013}
Stephen Roller and Sabine Schulte~im Walde. 2013.
\newblock A multimodal lda model integrating textual, cognitive and visual
  modalities.
\newblock In \emph{Proceedings of the 2013 Conference on Empirical Methods in
  Natural Language Processing}, pages 1146--1157.

\bibitem[{Socher et~al.(2013)Socher, Perelygin, Wu, Chuang, Manning, Ng, and
  Potts}]{Socher:2013}
Richard Socher, Alex Perelygin, Jean~Y. Wu, Jason Chuang, Christopher~D.
  Manning, Andrew~Y. Ng, and Christopher Potts. 2013.
\newblock Recursive deep models for semantic compositionality over a sentiment
  treebank.
\newblock In \emph{Proceedings of the 2013 conference on empirical methods in
  natural language processing}, pages 1631--1642.

\bibitem[{Tsvetkov et~al.(2016)Tsvetkov, Faruqui, and Dyer}]{Tsvetkov2016}
Yulia Tsvetkov, Manaal Faruqui, and Chris Dyer. 2016.
\newblock Correlation-based intrinsic evaluation of word vector
  representations.
\newblock In \emph{Proceedings of the 1st Workshop on Evaluating Vector-Space
  Representations for NLP}, pages 111--115.

\bibitem[{Tsvetkov et~al.(2015)Tsvetkov, Faruqui, Ling, Lample, and
  Dyer}]{Tsvetkov2015}
Yulia Tsvetkov, Manaal Faruqui, Wang Ling, Guillaume Lample, and Chris Dyer.
  2015.
\newblock Evaluation of word vector representations by subspace alignment.
\newblock In \emph{Proc. of EMNLP}.

\bibitem[{Turney(2012)}]{Turney2012}
Peter~D Turney. 2012.
\newblock Domain and function: A dual-space model of semantic relations and
  compositions.
\newblock \emph{Journal of Artificial Intelligence Research}, 44:533--585.

\bibitem[{Vinson and Vigliocco(2008)}]{Vinson2008}
David~P. Vinson and Gabriella Vigliocco. 2008.
\newblock Semantic feature production norms for a large set of objects and
  events.
\newblock \emph{Behavior Research Methods}, 40(1):183--190.

\bibitem[{Yaghoobzadeh and Sch{\"u}tze(2016)}]{Yaghoobzadeh2016}
Yadollah Yaghoobzadeh and Hinrich Sch{\"u}tze. 2016.
\newblock Intrinsic subspace evaluation of word embedding representations.
\newblock In \emph{Proceedings of the 54th Annual Meeting of the Association
  for Computational Linguistics (Volume 1: Long Papers)}, volume~1, pages
  236--246.

\bibitem[{Zhou and Xu(2015)}]{Zhou:2015}
Jie Zhou and Wei Xu. 2015.
\newblock End-to-end learning of semantic role labeling using recurrent neural
  networks.
\newblock In \emph{Proceedings of the 53rd Annual Meeting of the Association
  for Computational Linguistics and the 7th International Joint Conference on
  Natural Language Processing (Volume 1: Long Papers)}, volume~1, pages
  1127--1137.

\end{thebibliography}
\bibliographystyle{acl_natbib_nourl}

\end{document}